\documentclass{article} 
\usepackage{iclr2016_conference,times}
\usepackage{hyperref}
\usepackage{url}
\usepackage{amsfonts}
\usepackage{graphicx}
\usepackage{subfigure}
\usepackage[ruled]{algorithm2e}
\usepackage{bm}

\def\1{\bm{1}}

\def\vh{{\bm{h}}}

\def\vv{{\bm{v}}}

\def\vx{{\bm{x}}}
\def\vy{{\bm{y}}}

\def\vtheta{{\bm{\theta}}}

\def\mI{{\bm{I}}}

\def\mU{{\bm{U}}}

\def\mW{{\bm{W}}}

\DeclareMathAlphabet{\mathsfit}{\encodingdefault}{\sfdefault}{m}{sl}
\SetMathAlphabet{\mathsfit}{bold}{\encodingdefault}{\sfdefault}{bx}{n}

\def\sR{{\mathbb{R}}}

\usepackage{titlesec}
\titlespacing{\section}{0pt}{\parskip}{-\parskip}
\titlespacing{\subsection}{0pt}{\parskip}{-\parskip}
\titlespacing{\subsubsection}{0pt}{\parskip}{-\parskip}

\title{ {\tt Net2Net}: Accelerating Learning \\via Knowledge Transfer}

\author{Tianqi Chen\thanks{
Tianqi Chen is also a PhD student at University of Washington.
},\ \ Ian Goodfellow, \ and Jonathon Shlens
\\
Google Inc., Mountain View, CA\\
\texttt{tqchen@cs.washington.edu}, \texttt{\{goodfellow,shlens\}@google.com} \\
}

\iclrfinalcopy 

\begin{document}

\maketitle

\begin{abstract}
We introduce techniques for rapidly transferring the information stored in
one neural net into another neural net. The main purpose is to accelerate
the training of a significantly larger neural net.
During real-world workflows, one often trains very many different neural
networks during the experimentation and design process.
This is a wasteful process in which each new model is trained from scratch.
Our {\tt Net2Net} technique accelerates the experimentation process by
instantaneously transferring the knowledge from a previous network to each
new deeper or wider network.
Our techniques are based
on the concept of {\em function-preserving transformations} between neural
network specifications. This differs from previous approaches to
pre-training that altered the function represented by a neural net when
adding layers to it. Using our knowledge transfer mechanism to add depth to
Inception modules, we demonstrate a new state of the art accuracy rating on
the ImageNet dataset.
\end{abstract}

\section{Introduction}

We propose a new kind of operation to perform on large neural networks:
rapidly transfering knowledge contained in one neural network to another
neural network. We call this the {\tt Net2Net} family of operations.
We use {\tt Net2Net} as a general term describing any process
of training a {\em student} network significantly faster than would
otherwise be possible by leveraging knowledge from a {\em teacher} network
that was already trained on the same task.
In this article, we propose two specific {\tt Net2Net} methodologies. Both are
based on the idea of {\em function-preserving transformations} of neural networks.
Specifically, we initialize the student to be a neural network that
represents the same function as the teacher, but using a different
parameterization.
One of these transformations, {\tt Net2WiderNet} allows replacing a model with
an equivalent model that is wider (has more units in each hidden layer).
Another of these transformations, {\tt Net2DeeperNet} allows replacing a model
that satisfies some properties with an equivalent, deeper model.
After initializing the larger network to contain all of the knowledge
previously acquired by the smaller network, the larger network may be
trained to improve its performance.

Traditionally, machine learning algorithms have been designed to receive
a fixed dataset as input, initialize a new model with no knowledge, and
train that model to convergence on that dataset.
Real workflows are considerably more complicated than this idealized
scenario. We advocate {\tt Net2Net} operations as a useful tool for
accelerating real-world workflows.

One way that real workflows deviate from the idealized scenario is that
machine learning practitioners usually do not train only a single
model on each dataset. Instead, one typically trains multiple models, with
each model designed to improve upon the previous model in some way. Each
step in the iterative design process relies on fully training and evaluating
the innovation from the previous step. For many large models, training is
a long process, lasting for a week or even for a month. This makes
data-driven iterative design slow, due to the latency of evaluating whether
each change to the model caused an improvement.

\begin{figure}
\centering
\includegraphics[width=26pc]{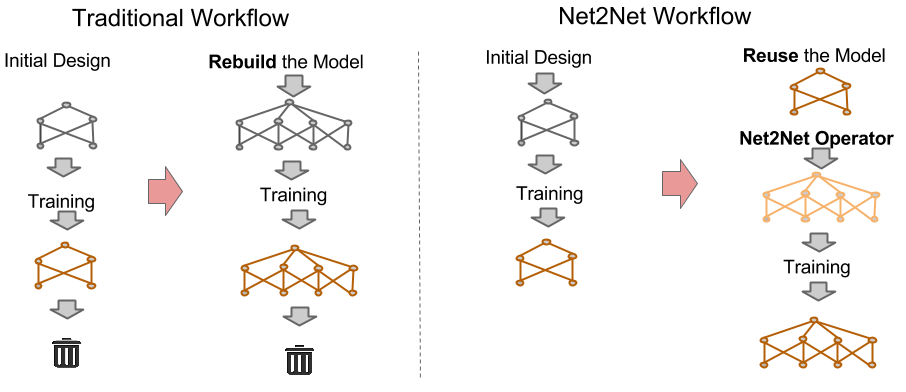}
\vspace{-.1in}
\caption{Comparison between a traditional workflow and the {\tt Net2Net} Workflow;
{\tt Net2Net} reuses information from an already trained model to speed up
the training of a new model.}
\label{fig:workflow}
\vspace{-.25in}
\end{figure}

{\tt Net2Net} operations accelerate these workflows by rapidly transferring
knowledge from the previous best model into each new model that an
experimenter proposes. Instead of training each considered design of model
for as much as a month, the experimenter can use {\tt Net2Net} to train the model
for a shorter period of time beginning from the function learned by the
previous best model. Fig~\ref{fig:workflow} demonstrates the difference of this approach
from traditional one.

More ambitiously, real machine learning systems will eventually become
lifelong learning systems~\citep{thrun95a,silver2013lifelong,NELL-aaai15}.
These machine learning systems need to continue to function for long periods
of time and continually experience new training examples as these examples
become available. We can think of a lifelong learning system as experiencing
a continually growing training set. The optimal model complexity changes
as training set size changes over time.
Initially, a small model may be preferred, in order to prevent overfitting and to reduce
the computational cost of using the model. Later, a large model may be
necessary to fully utilize the large dataset.
{\tt Net2Net} operations allow us to smoothly instantiate a significantly larger
model and immediately begin using it in our lifelong learning system, rather
than needing to spend weeks or months re-train a larger model from scratch on the latest,
largest version of the training set.

\section{Methodology}

In this section, we describe our new {\tt Net2Net} operations and how we
applied them on real deep neural nets.

\subsection{Feature prediction}

We briefly experimented with a method that proved not to offer a significant
advantage:
training a large student network beginning from a random initialization, and
introducing a set of extra ``teacher prediction'' layers into the student
network. Specifically, several convolutional hidden layers of the student
network were provided as input to new, learned, convolutional layers. The
cost function was modified to include terms encouraging the output of these
auxiliary layers to be close to a corresponding layer in the teacher network.
In other words, the student is trained to use each of its hidden layers to
predict the values of the hidden layers in the teacher.

The goal was that the teacher would provide a good internal representation
for the task that the student could quickly copy and then begin to refine.
The approach resembles the FitNets~\citep{Romero-et-al-TR2014} strategy for
training very thin networks of moderate depth.
Unfortunately, we did not find that this method offered any compelling
speedup or other advantage relative to the baseline approach.
This may be because our baseline was very strong, based on training with
batch normalization~\citep{Ioffe+Szegedy-2015}.
\citet{mahayri2015} independently observed that the benefits of the FitNets
training strategy were eliminated after changing the model to use batch
normalization.

The FitNets-style approach to {\tt Net2Net} learning is very general, in the sense that,
if successful, it
would allow any architecture of student network to learn from any architecture of
teacher network. Though we were not able to make this general approach work,
we encourage other researchers to attempt to design fully general
{\tt Net2Net} strategies in the future.
We instead turned to different {\tt Net2Net} strategies that were limited in
scope but more effective.

\subsection{Function-preserving initializations}

We introduce two effective {\tt Net2Net} strategies. Both are based on
initializing the student network to represent the same function as the
teacher, then continuing to train the student network by normal means.
Specifically, suppose that a teacher network is represented by a function
$ \vy = f(\vx ; \vtheta) $
where $\vx$ is the input to the network, $\vy$ is the output of the network,
and $\vtheta$ is the parameters of the network.
In our experiments, $f$ is defined by a convolutional network,
$\vx$ is an input image, and $\vy$ is a vector of probabilities representing
the conditional distribution over object categories given $\vx$.

Our strategy is to choose a new set of parameters $\vtheta'$ for a student
network $g(\vx ; \vtheta')$ such that
\[ \forall \vx, f(\vx; \vtheta) = g(\vx; \vtheta') .\]

In our description of the method, we assume for simplicity that both the teacher
and student networks contain compositions of
standard linear neural network layers of the form
$\vh^{(i)} = \phi( \mW^{(i)\top} \vh^{(i-1)} )$
where $\vh^{(i)}$ is the activations of hidden layer $i$, $\mW^{(i)}$ is the
matrix of incoming weights for this layer, and $\phi$ is an activation function,
such as as the rectified linear activation function~\citep{Jarrett-ICCV2009,Glorot+al-AI-2011-small}
or the maxout activation function~\citep{Goodfellow_maxout_2013}.
We use this layer formalization for clarity, but our method generalizes to arbitrary
composition structure of these layers, such as Inception~\citep{Szegedy-et-al-arxiv2014}.
All of our experiments are in fact performed using Inception networks.

Though we describe the method in terms of matrix multiplication for
simplicity, it readily extends to convolution (by observing that
convolution is multiplication by a doubly block circulant matrix)
and to the addition of biases (by treating biases as a row of
$\mW$ that are multiplied by a constant input of $1$).
We will also give a specific description for the convolutional
case in subsequent discussions of the details of the method.
Function-preserving initialization carries many advantages over
other approaches to growing the network:
\begin{itemize}
\item The new, larger network immediately performs as well as the
original network, rather than spending time passing through
a period of low performance.
\item Any change made to the network after initialization is guaranteed
to be an improvement, so long as each local step is an improvement.
Previous methods could fail to improve over the baseline even if
each local step was guaranteed to be an improvement, because the
initial change to the larger model size could worsen performance.
\item It is always ``safe'' to optimize all parameters in the network.
There is never a stage where some layers receive harmful gradients
and must be frozen. This is in contrast to approaches such as cascade
correlation, where the old units are frozen in order to avoid making
poor adaptations while attempting to influence the behavior of new
randomly connected units.
\end{itemize}

\subsection{Net2WiderNet}
\begin{figure}
\centering
\includegraphics{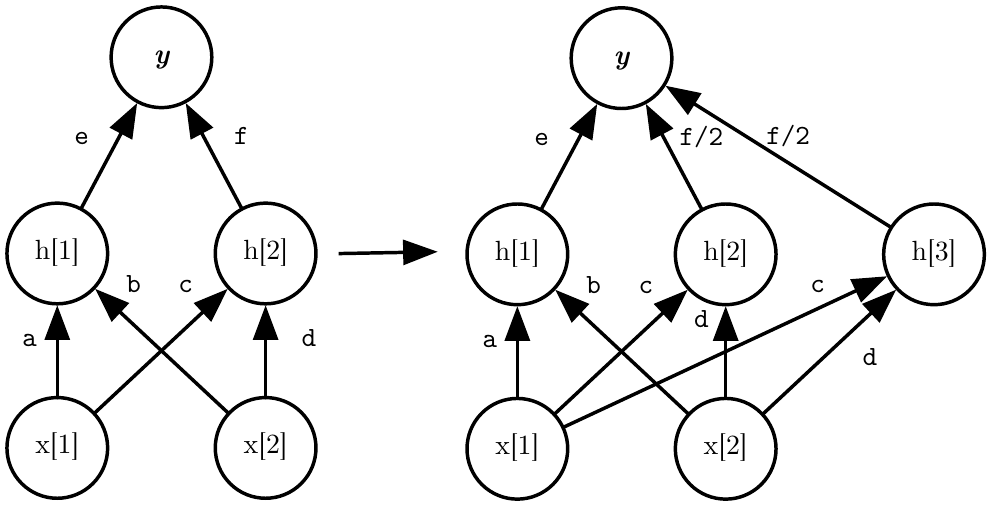} 
\vspace{-.1in}
\caption{The {\tt Net2WiderNet} transformation.
In this example, the teacher network has an input layer with
two inputs $x[1]$ and $x[2]$, a hidden layer with two rectified
linear hidden units $h[1]$ and $h[2]$, and an output $y$.
We use the {\tt Net2WiderNet} operator to create a student network
that represents the same function as the teacher network.
The student network is larger because we replicate the $h[2]$ unit
of the teacher.
The labels on the edges indicate the value of the associated weights.
To replicate the $h[2]$ unit, we copy its weights $c$ and $d$ to the
new $h[3]$ unit.
The weight $f$, going out of $h[2]$, must be copied to also go out
of $h[3]$. This outgoing weight must also be divided by $2$ to
compensate for the replication of $h[2]$.
This is a simple example intended to illustrate the conceptual
idea. For a practical application, we would simultaneously replicate
many randomly chosen units, and we would add a small amount of noise
to break symmetry after the replication.
We also typically widen many layers rather than just one layer,
by recursively applying the {\tt Net2WiderNet} operator.
}\label{fig:net2widernet}
\vspace{-.2in}
\end{figure}
\begin{algorithm}[t]
    \caption{Net2WiderNet}\label{alg:net2widernet}
    \KwIn{$\{\mW^{(i)}| i = 1,2, \cdots n\}$, the weight matrix of teacher net}
    Use forward inference to generate a consistent random mapping $\{g^{(i)}\}$\\
    \For{$i \in 1, 2 , \cdots n$} {
        $c_j \leftarrow 0 $
        \For{$j \in 1, 2, \cdots q$} {
            $c_{g^{(i-1)}(j)} \leftarrow c_{g^{(i-1)}(j)} + 1$
        }
        \For{$j \in 1, 2, \cdots q$} {
            $\mU^{(i)}_{k,j} \leftarrow \frac{1}{c_j} \mW^{(i)}_{g^{(i-1)}(k), g^{(i)}(j)}$
        }
    }
    \KwOut{\{$\mU^{(i)}| i = 1,2, \cdots n\}$: the transformed weight matrix for wider net.}
\end{algorithm}

Our first proposed transformation is {\tt Net2WiderNet} transformation.
This allows a layer to be replaced with a wider layer, meaning a layer that
has more units. For convolution architectures, this means the layers will have more convolution
channels.

Let us start with a simple special case to illustrate the operation.
Suppose that layer $i$ and layer $i+1$ are both fully connected layers, and
layer $i$ uses an elementwise non-linearity. To widen layer $i$, we replace $\mW^{(i)}$ and $\mW^{(i+1)}$.
If layer $i$ has $m$ inputs and $n$ outputs, and layer $i+1$ has $p$ outputs, then $\mW^{(i)} \in \sR^{m \times n}$ and
$\mW^{(i+1)} \in \sR^{n \times p}$.
{\tt Net2WiderNet} allows us to replace layer $i$ with a layer that has $q$ outputs, with $q > n$.
We will introduce a random mapping function $g: \{1,2, \cdots, q\} \rightarrow \{1,2, \cdots, n\}$, that satisfies
\[
g(j) = \left\{
\begin{array}{ll}
j & j \leq n\\
\mbox{ random sample from } \{1,2,\cdots n\} & j > n\\
\end{array}
\right.
\]

We introduce a new weight matrix $\mU^{(i)}$ and $\mU^{(i+1)}$ representing
the weights for these layers in the new student network. Then the
new weights are given by
\[
	\mU^{(i)}_{k,j} =  \mW^{(i)}_{k, g(j)}, \ \ \ \mU^{(i+1)}_{j,h} = \frac{1}{|\{x| g(x) = g(j)\}|} \mW^{(i+1)}_{g(j), h}
    .
\]

Here, the first $n$ columns of $\mW^{(i)}$ are copied directly into $\mU^{(i)}$.
Columns $n+1$ through $q$ of $\mU^{{(i)}}$ are created by choosing a random as defined in $g$.
The random selection is performed with replacement, so each column of $\mW^{(i)}$ is copied potentially many times.
For weights in $\mU^{(i+1)}$, we must account for the replication by dividing the weight by replication
factor given by $\frac{1}{|\{x| g(x) = g(j)\}|}$, so all the units have the exactly the same value as the unit in the original net.

This description can be generalized to making multiple layers wider, with the layers composed
as described by an arbitrary directed acyclic computation graph.
This general procedure is illustrated by Fig.~\ref{fig:net2widernet}.
So far we have only discussed the use of a single random mapping function to expand one layer.
We can in fact introduce a random mapping function $g^{(i)}$ for every non-output layer.
Importantly, these $g^{(i)}$ are subject to some constraints as defined by the computation graph.
Care needs to be taken to ensure that the remapping functions do in fact result in function
preservation.

To explain, we provide examples of two computational graph structures that impose specific constraints
on the random remapping functions.

One example is the layer structure used by batch normalization~\citep{Ioffe+Szegedy-2015}.
The layer involves both a standard linear transformation, but also involves elementwise
multiplication by learned parameters that allow the layer to represent any range of outputs
despite the normalization operation.
The random remapping for the multiplication parameters must match the random remapping for
the weight matrix. Otherwise we could generate a new unit that uses the weight vector for
pre-existing unit $i$ but is scaled by the multiplication parameter for unit $j$. The new
unit would not implement the same function as the old unit $i$ or as the old unit $j$.

Another example is concatenation. If we concatenate the output of layer 1 and layer 2, then
pass this concatenated output to layer 3, the remapping function for layer 3 needs to take
the concatenation into account. The width of the output of layer 1 will determine the offset
of the coordinates of units originating in layer 2.

To make {\tt Net2WiderNet} a fully general algorithm, we would need a \emph{remapping inference} algorithm
that makes a forward pass through the graph, querying each operation in the graph about how to
make the remapping functions consistent. For our experiments, we manually designed all of the inference necessary rules
for the Inception network, which also works for most feed forward network.
This is similar to the existing shape inference functionality that allows us to predict the shape
of any tensor in the computational graph by making a forward pass through the graph, beginning
with input tensors of known shape.

After we get the random mapping, we can copy the weight over and divide by the replication factor, which is formally given
by the following equation.
\[
	\mU^{(i)}_{k,j} = \frac{1}{|\{x| g^{(i-1)}(x) = g^{(i-1)}(k)\}|} \mW^{(i)}_{g^{(i-1)}(k), g^{(i)}(j)}
\]

It is essential that each unit be replicated at least once, hence the
constraint that the resulting layer be wider.
This operator can be applied arbitrarily many times; we can expand only
one layer of the network, or we can expand all non-output layers.

In the setting where several units need to share the same weights, for example convolution operations.
We can add such constraint to the random mapping generation, such that source of weight is consistent.
This corresponds to make a random mapping on the channel level, instead of unit level, the rest
procedure remains the same.

When training with certain forms of randomization on the widened layer,
such as dropout \citep{Srivastava14},
it is acceptable to use perfectly transformation preserving {\tt Net2WiderNet}, as we have described so far.
When using a training algorithm that does not
use randomization to encourage identical units to learn different functions,
one should add a small amount of noise to all but the first copy of each
column of weights. This results in the student network representing only
approximately the same function as the teacher, but this approximation is
necessary to ensure that the student can learn to use its full capacity when
training resumes.

\subsection{Net2DeeperNet}
\begin{figure}
\centering
\includegraphics[width=.7\textwidth]{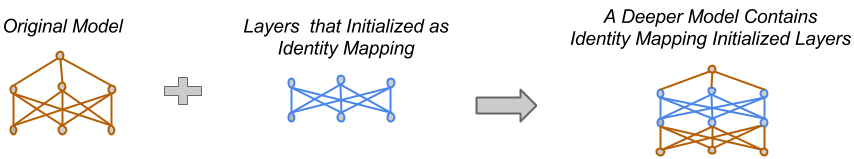}
\vspace{-.12in}
\caption{The Net2DeeperNet Transformation}\label{fig:net2deepernet}
\vspace{-.2in}
\end{figure}
We also introduce a second function-preserving transformation,
{\tt Net2DeeperNet}.
This allows us to transform an existing net into a deeper one.
Specifically, the {\tt Net2DeeperNet} replaces a layer
$\vh^{(i)} = \phi(\vh^{(i-1)\top} \mW^{(i)})$
with two layers
$\vh^{(i)} = \phi\left(
		\mU^{(i)\top} \phi\left(
			\mW^{(i)\top} \vh^{(i-1)}
\right)
\right). $
The new matrix $\mU$ is initialized to an identity matrix, but remains free
to learn to take on any value later.
This operation is only applicable when $\phi$ is chosen such that
$ \phi(\mI \phi( \vv )) = \phi( \vv) $
for all vectors $\vv$.
This property holds for the rectified linear activation.
To obtain {\tt Net2DeeperNet} for maxout units, one must use a matrix that
is similar to identity, but with replicated columns.
Unfortunately, for some popular activation functions, such as the logistic
sigmoid, it is not possible to insert a layer of the same type that
represents an identity function over the required domain.
When we apply it to convolution networks, we can simply set
the convolution kernels to be identity filters.

In some cases, to build an identity layer requires additional work.
For example, when using batch normalization, we must set the output
scale and output bias of the normalization layer to undo the normalization
of the layer's statistics. This requires running forward propagation
through the network on training data in order to estimate the activation
statistics.

The approach we take is a specific case of a more general approach, that is to
build a multiple layer network that factorizes the original layer. Making
a deeper but equivalent representation. However it is hard to do general factorization
of layers which non-linear transformation units, such as rectified linear and batch normalization.
Restricting to adding identity transformation allows us to handle such non-linear transformations,
and apply the methodology to the network architectures that we care about. It also
saves the computation cost of doing such factorization. We think support for general factorization
of layers is an interesting future direction that is worth exploring.

When using {\tt Net2DeeperNet}, there is no need to add noise. We can begin
training with a student network that represents exactly the same function
as the teacher network. Nevertheless, a small amount of noise can be added to break the symmetry more rapidly.
At first glance, it appears that {\tt Net2Net} can only add a new layer
with the same width as the layer below it, due to the use of the identity
operation. However, {\tt Net2WiderNet} may be composed with {\tt Net2DeeperNet},
so we may in fact add any hidden layer that is at least as wide as the
layer below it.

\section{Experiments}
\subsection{Experimental Setup}
We evaluated our {\tt Net2Net} operators in three different settings.
In all cases we used an Inception-BN network~\citep{Ioffe+Szegedy-2015} trained on ImageNet.
In the first setting, we demonstrate that {\tt Net2WiderNet}
can be used to accelerate the training of a standard Inception
network by initializing it with a smaller network.
In the second setting, we demonstrate that {\tt Net2DeeperNet} allows us
to increase the depth of the Inception modules.
Finally, we use our {\tt Net2Net} operators in a realistic setting,
where we make more dramatic changes to the model size and explore
the model space for better performance. In this setting, we demonstrate
an improved result on ImageNet.

We will be comparing our method to some baseline methods:
\begin{itemize}
	\item ``Random pad'': This is a baseline for {\tt Net2WiderNet}.
		We widen the network by adding new units with random weights,
		rather than by replicating units to perform function-preserving
		initialization. This operation is implemented by padding the
        pre-existing weight matrices with additional random values.
    \item ``Random initialization'' : This is a baseline for {\tt Net2DeeperNet}.
        We compare the training of a deep network accelerated by initialization
        from a shallow one against the training of an identical deep network
        initialized randomly.
\end{itemize}

\subsection{Net2WiderNet}

\begin{figure*}
\vspace{-.24in}
  \centering
  \subfigure[Training Accuracy of Different Methods]{
    \includegraphics[width=26pc]{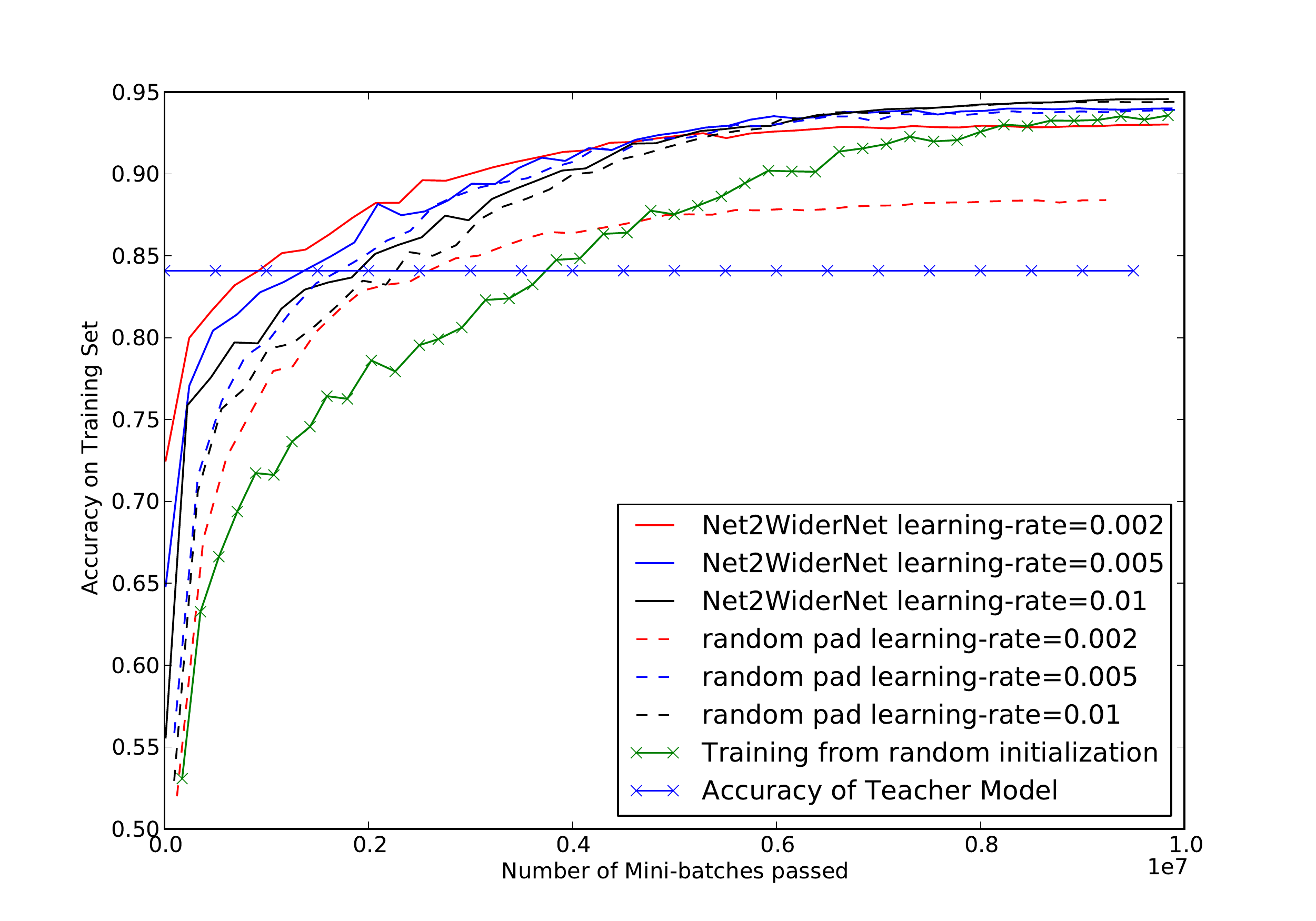}
  }
  \subfigure[Validation Accuracy of Different Methods]{
    \includegraphics[width=26pc]{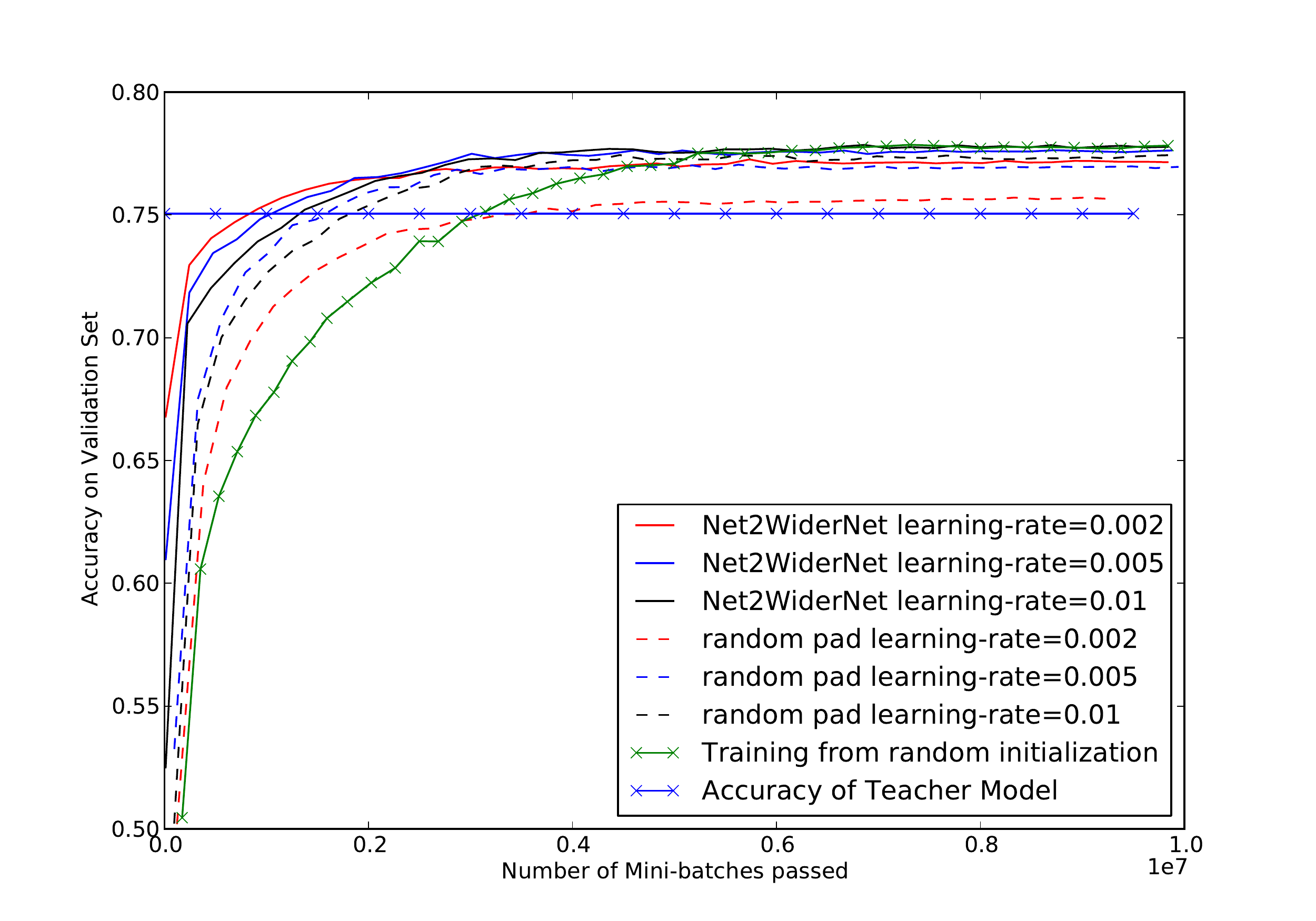}
  }
  \caption{Comparison of various approaches for training a wider model.
  {\tt Net2Net} provides a model that is useful almost immediately
  and reaches high levels of performance faster than the baseline approaches.
  In particular, {\tt Net2WiderNet} converges to roughly its final
  validation set accuracy after roughly $3 \times 10^6$ minibatches.
  The randomly initialized baseline converges to the same validation
  set accuracy but requires roughly an additional $2 \times 10^6$
  minibatches to do so.
  }
  \label{fig:small2standard}
\vspace{-.24in}
\end{figure*}

We start by evaluating the method of {\tt Net2WiderNet}. We began by constructing a teacher network that was narrower than the
standard Inception. We reduced the number of convolution channels at each layer within all
Inception modules by a factor of $\sqrt{0.3}$.
Because both the input and output number of channels are reduced, this
reduces the number of parameters in most layers to 30\% of the original amount.
To simplify the software for our experiments, we did not modify any
component of the network other than the Inception modules.
After training this small teacher network, we used it to accelerated the
training of a standard-sized student network.

Fig. \ref{fig:small2standard} shows the comparison of different approaches. We can find that
the proposed approach gives faster convergence than the baseline approaches. Importantly,
{\tt Net2WiderNet} gives the same level of final accuracy as the model trained from random initialization.
This indicates that the true size of the model governs the accuracy that the training
process can attain.
There is no loss in accuracy that comes from initializing the model to mimic
a smaller one.
{\tt Net2WiderNet}
can thus be safely used to get to the same accuracy quicker,
reducing the time required to run new experiments.

\subsection{Net2DeeperNet}
\begin{figure*}
  \centering
  \subfigure[Training Accuracy of Different Methods]{
    \includegraphics[width=26pc]{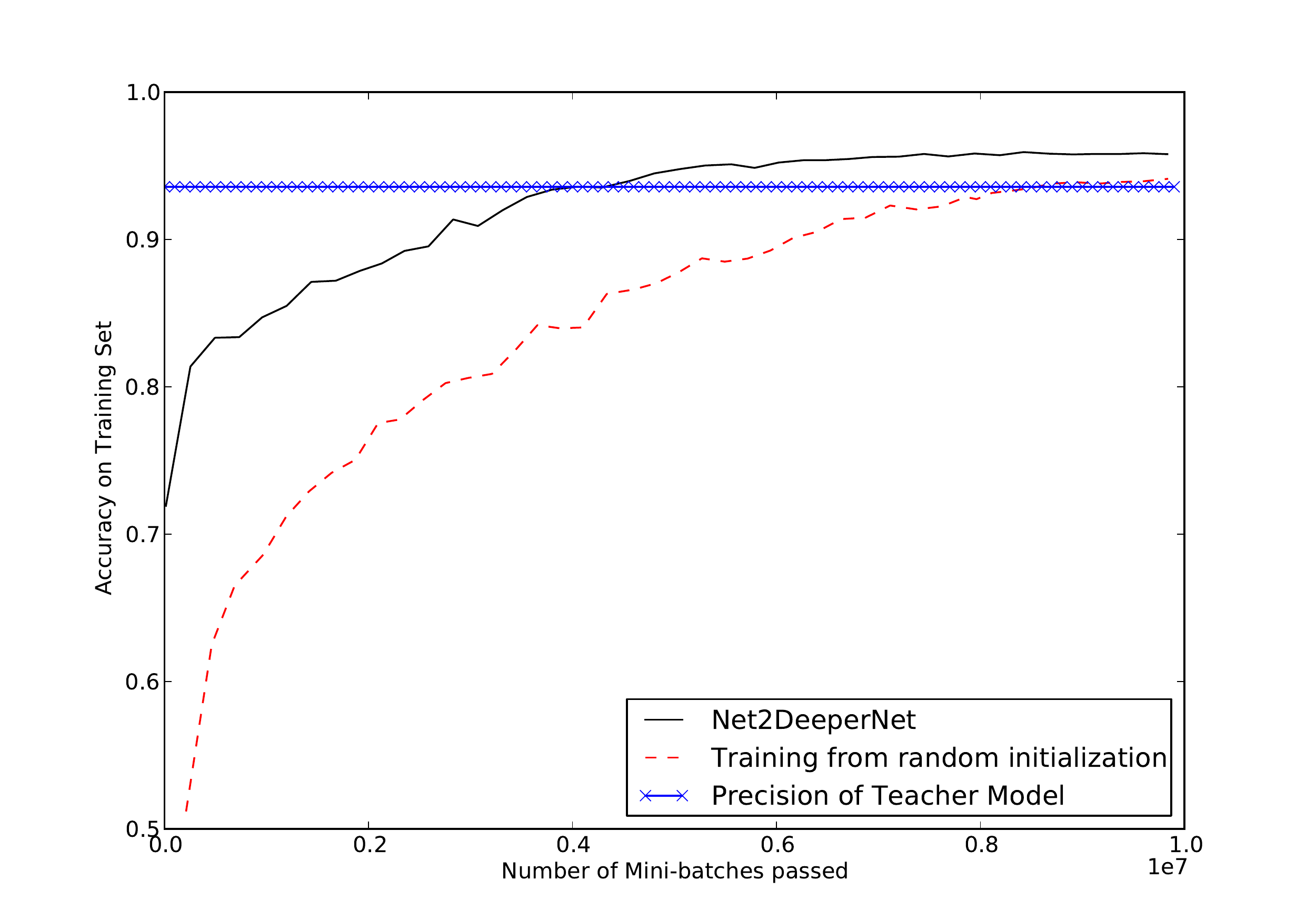}
  }
  \subfigure[Validation Accuracy of Different Methods]{
    \includegraphics[width=26pc]{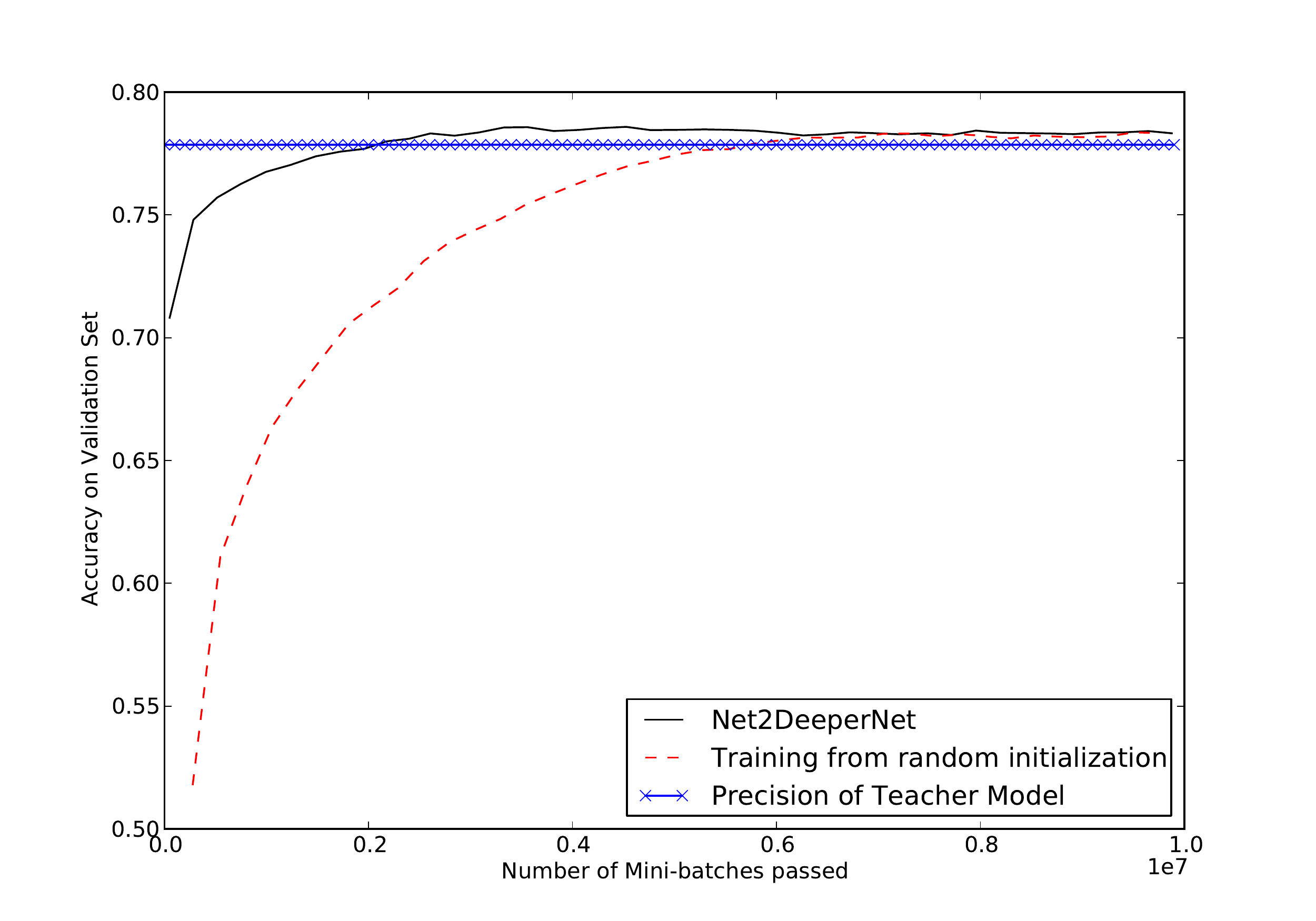}
  }
    \vspace{-.1in}
  \caption{Comparison of methods of training a deeper model}
  \label{fig:deeper}
  \vspace{-.3in}
\end{figure*}

We conducted experiments with using {\tt Net2DeeperNet}
to make the network deeper.
For these experiments, we used a standard Inception model as the teacher network,
and increased the depth of each inception module.
The convolutional layers in Inception modules use rectangular kernels.
The convolutional layers are arranged in pairs, with a layer using a vertical
kernel followed by a layer using a horizontal kernel. Each of these layers
is a complete layer with a rectifying non-linearity and batch normalization;
it is not just a factorization of a linear convolution operation into separable
parts.
Everywhere that a pair of
vertical-horizontal convolution layers appears, we added two more pairs of such
layers, configured to result in an identity transformation.
The results are shown in Fig.~\ref{fig:deeper}. We find that {\tt Net2DeeperNet}
obtains good
accuracy much faster than training from random initialization, both in terms of
training and validation accuracy.

\subsection{Exploring Model Design Space with {\tt Net2Net}}
\begin{figure*}
  \centering
  \subfigure[Training Accuracy of Different Methods]{
    \includegraphics[width=26pc]{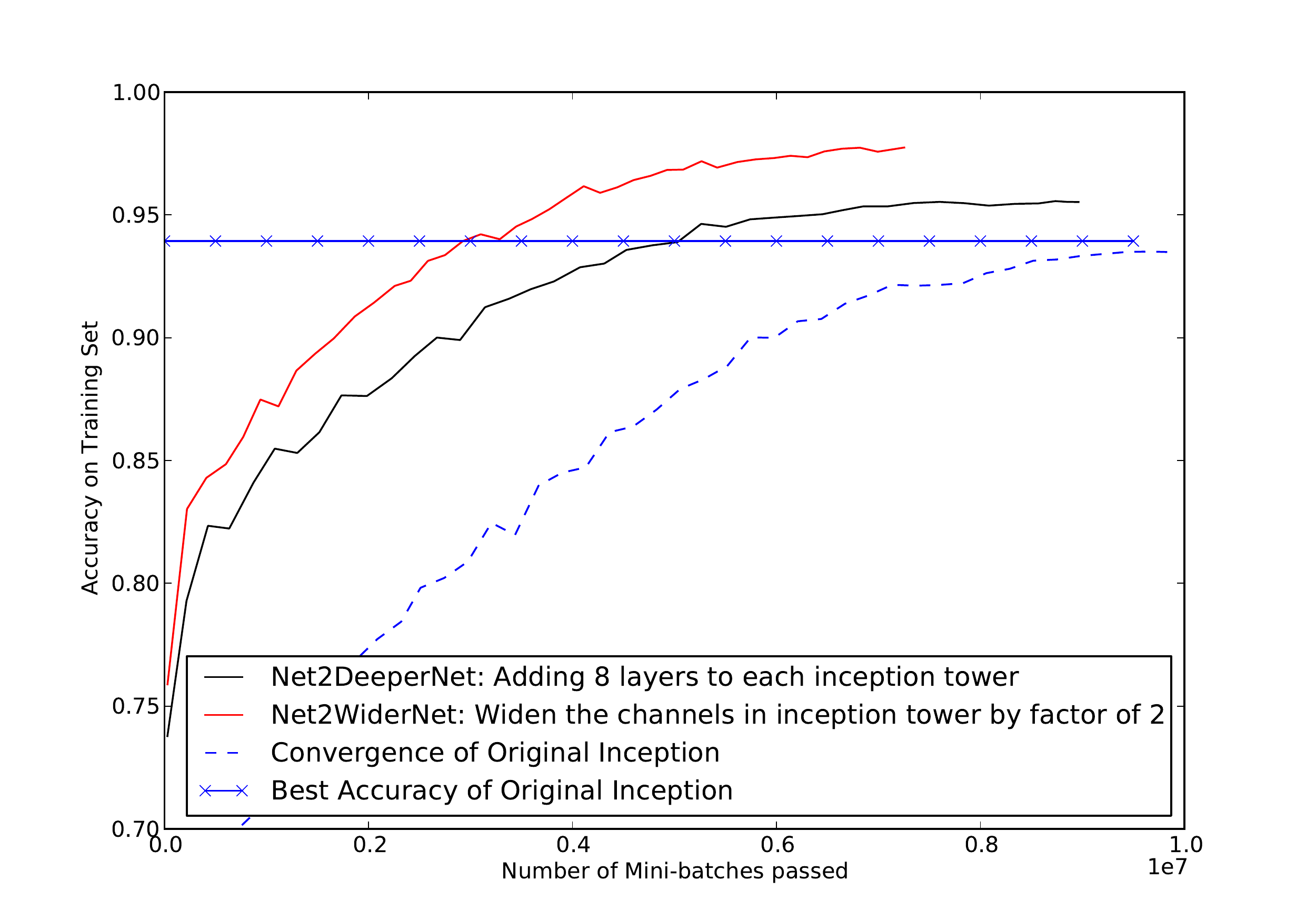}
  }
  \subfigure[Validation Accuracy of Different Methods]{
    \includegraphics[width=26pc]{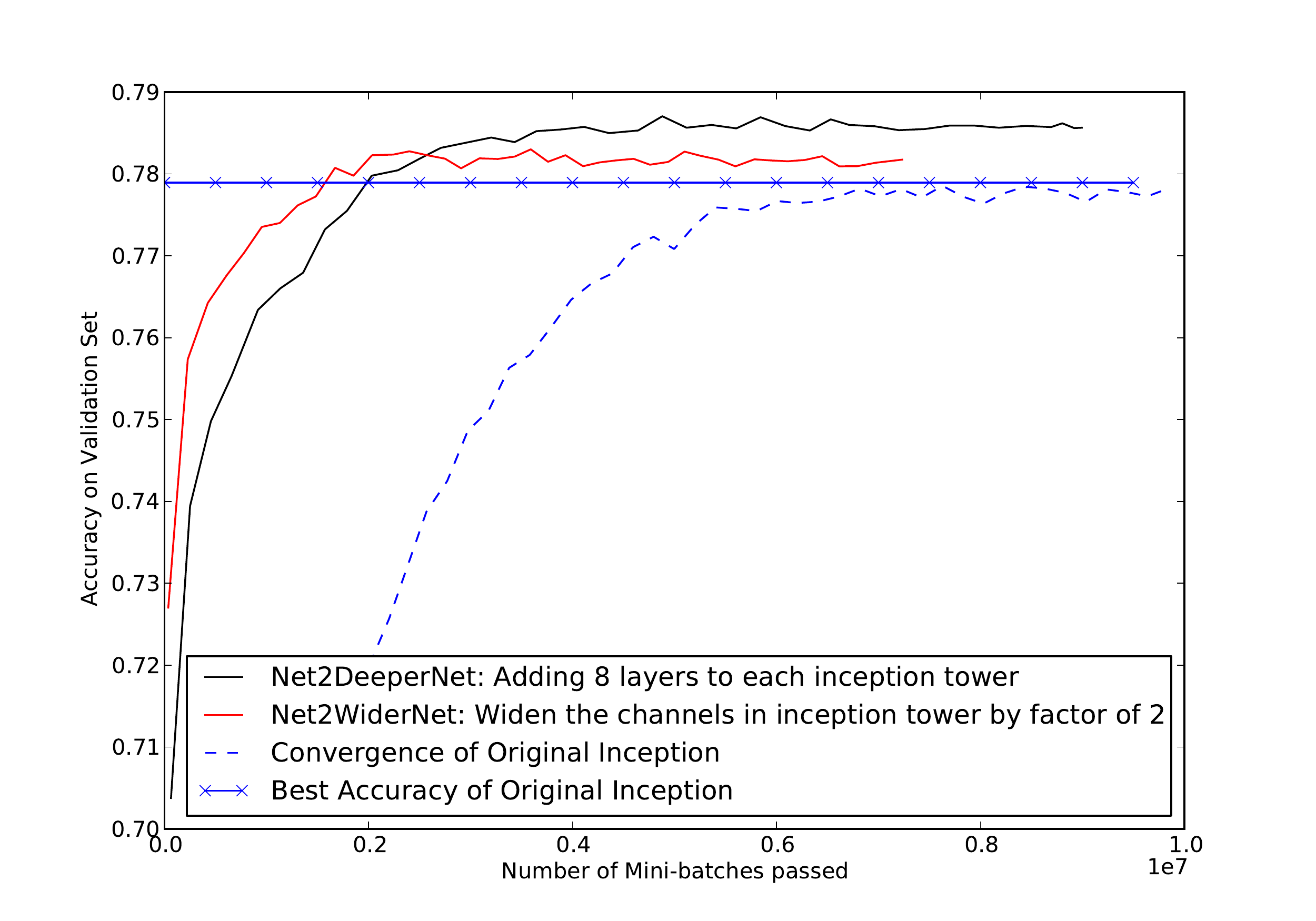}
  }
      \vspace{-.1in}
  \caption{Using {\tt Net2Net} to quickly explore designs of larger nets.
  By adding eight layers to each Inception module, we obtained a new
  state of the art test error on ImageNet.
  }
  \label{fig:explore}
\vspace{-.26in}
\end{figure*}

One of important property of {\tt Net2Net} is that it enables quick exploration
of modelp space, by transforming an existing state-of-art architecture.
In this experiment, we made an ambitious exploration of model design space in both wider and deeper directions.
Specifically, we enlarged the width of an Inception model to $\sqrt{2}$ times of the original one.
We also built another deeper net by adding four vertical-horizontal convolutional layer pairs on top of every inception modules in the original Inception model.

The results are shown in Fig. \ref{fig:explore}. This last approach paid off, yielding a model that sets a new state of the
art of 78.5\% on our ImageNet validation set.
We did not train these larger models from scratch, due to resource and time constraints.
However,  we reported the convergence curve of  the original inception model for reference, which should be easier to train than these larger models.
We can find that the models initialized with {\tt Net2Net} operations converge even faster than the standard model.
This example really demonstrate the advantage of {\tt Net2Net} approach which helps us to explore the design space faster and advance the results in deep learning.

\section{Discussion}

Our {\tt Net2Net} operators have demonstrated that it is possible
to rapidly transfer knowledge from a small neural network to a
significantly larger neural network under some architectural
constraints.
We have demonstrated that we can train larger neural networks
to improve performance on ImageNet recognition using this
approach.
{\tt Net2Net} may also now be used as a technique for exploring
model families more rapidly, reducing the amount of time needed
for typical machine learning workflows.
We hope that future research will uncover new ways of transferring
knowledge between neural networks.
In particular, we hope future research will reveal more general
knowledge transfer methods that can rapidly initialize a student
network whose architecture is not constrained to resemble that
of the teacher network.

\subsubsection*{Acknowledgments}

We would like to thank Jeff Dean and George Dahl for helpful discussions.
We also thank the developers of TensorFlow \citep{tensorflow2015-whitepaper},
which we used for all of our experiments. We would like to thank
Conrado Miranda for helpful feedback that we used to improve the
clarity and comprehensiveness of this manuscript.

\bibliographystyle{iclr2016_conference}
\small{
\bibliography{ml,aigaion}

\begin{thebibliography}{20}
\providecommand{\natexlab}[1]{#1}
\providecommand{\url}[1]{\texttt{#1}}
\expandafter\ifx\csname urlstyle\endcsname\relax
  \providecommand{\doi}[1]{doi: #1}\else
  \providecommand{\doi}{doi: \begingroup \urlstyle{rm}\Url}\fi

\bibitem[Abadi et~al.(2015)Abadi, Agarwal, Barham, Brevdo, Chen, Citro,
  Corrado, Davis, Dean, Devin, Ghemawat, Goodfellow, Harp, Irving, Isard, Jia,
  Jozefowicz, Kaiser, Kudlur, Levenberg, Man\'{e}, Monga, Moore, Murray, Olah,
  Schuster, Shlens, Steiner, Sutskever, Talwar, Tucker, Vanhoucke, Vasudevan,
  Vi\'{e}gas, Vinyals, Warden, Wattenberg, Wicke, Yu, and
  Zheng]{tensorflow2015-whitepaper}
Abadi, Mart\'{\i}n, Agarwal, Ashish, Barham, Paul, Brevdo, Eugene, Chen,
  Zhifeng, Citro, Craig, Corrado, Greg~S., Davis, Andy, Dean, Jeffrey, Devin,
  Matthieu, Ghemawat, Sanjay, Goodfellow, Ian, Harp, Andrew, Irving, Geoffrey,
  Isard, Michael, Jia, Yangqing, Jozefowicz, Rafal, Kaiser, Lukasz, Kudlur,
  Manjunath, Levenberg, Josh, Man\'{e}, Dan, Monga, Rajat, Moore, Sherry,
  Murray, Derek, Olah, Chris, Schuster, Mike, Shlens, Jonathon, Steiner,
  Benoit, Sutskever, Ilya, Talwar, Kunal, Tucker, Paul, Vanhoucke, Vincent,
  Vasudevan, Vijay, Vi\'{e}gas, Fernanda, Vinyals, Oriol, Warden, Pete,
  Wattenberg, Martin, Wicke, Martin, Yu, Yuan, and Zheng, Xiaoqiang.
\newblock {TensorFlow}: Large-scale machine learning on heterogeneous systems,
  2015.
\newblock URL \url{http://tensorflow.org/}.
\newblock Software available from tensorflow.org.

\bibitem[Bucilu\v{a} et~al.(2006)Bucilu\v{a}, Caruana, and
  Niculescu-Mizil]{bucilua2006model}
Bucilu\v{a}, Cristian, Caruana, Rich, and Niculescu-Mizil, Alexandru.
\newblock Model compression.
\newblock In \emph{Proceedings of the 12th ACM SIGKDD international conference
  on Knowledge discovery and data mining}, pp.\  535--541. ACM, 2006.

\bibitem[Fahlman \& Lebiere(1990)Fahlman and Lebiere]{Fahlman90}
Fahlman, Scott~E. and Lebiere, Christian.
\newblock The cascade-correlation learning architecture.
\newblock pp.\  524--532, Denver, CO, 1990. Morgan Kaufmann, San Mateo.

\bibitem[Glorot et~al.(2011)Glorot, Bordes, and
  Bengio]{Glorot+al-AI-2011-small}
Glorot, X., Bordes, A., and Bengio, Y.
\newblock Deep sparse rectifier neural networks.
\newblock In \emph{AISTATS'2011}, 2011.

\bibitem[Goodfellow et~al.(2013)Goodfellow, Warde-Farley, Mirza, Courville, and
  Bengio]{Goodfellow_maxout_2013}
Goodfellow, Ian~J., Warde-Farley, David, Mirza, Mehdi, Courville, Aaron, and
  Bengio, Yoshua.
\newblock Maxout networks.
\newblock In \emph{ICML'2013}, 2013.

\bibitem[Gutstein et~al.(2008)Gutstein, Fuentes, and
  Freudenthal]{gutstein2008knowledge}
Gutstein, Steven, Fuentes, Olac, and Freudenthal, Eric.
\newblock Knowledge transfer in deep convolutional neural nets.
\newblock \emph{International Journal on Artificial Intelligence Tools},
  17\penalty0 (03):\penalty0 555--567, 2008.

\bibitem[Hinton et~al.(2015)Hinton, Vinyals, and Dean]{hinton2015distilling}
Hinton, Geoffrey, Vinyals, Oriol, and Dean, Jeff.
\newblock Distilling the knowledge in a neural network.
\newblock \emph{arXiv preprint arXiv:1503.02531}, 2015.

\bibitem[Hinton et~al.(2006)Hinton, Osindero, and Teh]{Hinton06}
Hinton, Geoffrey~E., Osindero, Simon, and Teh, {Yee Whye}.
\newblock A fast learning algorithm for deep belief nets.
\newblock \emph{Neural Computation}, 18:\penalty0 1527--1554, 2006.

\bibitem[Ioffe \& Szegedy(2015)Ioffe and Szegedy]{Ioffe+Szegedy-2015}
Ioffe, Sergey and Szegedy, Christian.
\newblock Batch normalization: Accelerating deep network training by reducing
  internal covariate shift.
\newblock 2015.

\bibitem[Jarrett et~al.(2009)Jarrett, Kavukcuoglu, Ranzato, and
  {LeCun}]{Jarrett-ICCV2009}
Jarrett, Kevin, Kavukcuoglu, Koray, Ranzato, {Marc'Aurelio}, and {LeCun}, Yann.
\newblock What is the best multi-stage architecture for object recognition?
\newblock In \emph{Proc. International Conference on Computer Vision
  (ICCV'09)}, pp.\  2146--2153. IEEE, 2009.

\bibitem[Mahayri et~al.(2015)Mahayri, Ballas, and Courville]{mahayri2015}
Mahayri, Amjad, Ballas, Nicolas, and Courville, Aaron.
\newblock {FitNets} and batch normalization.
\newblock Technical report, (unpublished), 2015.

\bibitem[Mitchell et~al.(2015)Mitchell, Cohen, Hruschka, Talukdar, Betteridge,
  Carlson, Dalvi, Gardner, Kisiel, Krishnamurthy, Lao, Mazaitis, Mohamed,
  Nakashole, Platanios, Ritter, Samadi, Settles, Wang, Wijaya, Gupta, Chen,
  Saparov, Greaves, and Welling]{NELL-aaai15}
Mitchell, T., Cohen, W., Hruschka, E., Talukdar, P., Betteridge, J., Carlson,
  A., Dalvi, B., Gardner, M., Kisiel, B., Krishnamurthy, J., Lao, N., Mazaitis,
  K., Mohamed, T., Nakashole, N., Platanios, E., Ritter, A., Samadi, M.,
  Settles, B., Wang, R., Wijaya, D., Gupta, A., Chen, X., Saparov, A., Greaves,
  M., and Welling, J.
\newblock Never-ending learning.
\newblock In \emph{Proceedings of the Twenty-Ninth AAAI Conference on
  Artificial Intelligence (AAAI-15)}, 2015.

\bibitem[Romero et~al.(2014)Romero, Ballas, Ebrahimi~Kahou, Chassang, Gatta,
  and Bengio]{Romero-et-al-TR2014}
Romero, Adriana, Ballas, Nicolas, Ebrahimi~Kahou, Samira, Chassang, Antoine,
  Gatta, Carlo, and Bengio, Yoshua.
\newblock {FitNets}: Hints for thin deep nets.
\newblock Technical Report Arxiv report 1412.6550, arXiv, 2014.

\bibitem[Silver et~al.(2013)Silver, Yang, and Li]{silver2013lifelong}
Silver, DL, Yang, Q, and Li, L.
\newblock Lifelong machine learning systems: Beyond learning algorithms.
\newblock In \emph{AAAI Spring Symposium-Technical Report}, 2013.

\bibitem[Simonyan \& Zisserman(2015)Simonyan and Zisserman]{Simonyan2015}
Simonyan, Karen and Zisserman, Andrew.
\newblock Very deep convolutional networks for large-scale image recognition.
\newblock In \emph{ICLR}, 2015.

\bibitem[Socher et~al.(2013)Socher, Bauer, Manning, and
  Ng]{Socher13parsingwith}
Socher, Richard, Bauer, John, Manning, Christopher~D., and Ng, Andrew~Y.
\newblock Parsing with compositional vector grammars.
\newblock In \emph{In Proceedings of the ACL conference}, 2013.

\bibitem[Srivastava et~al.(2014)Srivastava, Hinton, Krizhevsky, Sutskever, and
  Salakhutdinov]{Srivastava14}
Srivastava, Nitish, Hinton, Geoffrey, Krizhevsky, Alex, Sutskever, Ilya, and
  Salakhutdinov, Ruslan.
\newblock Dropout: A simple way to prevent neural networks from overfitting.
\newblock \emph{Journal of Machine Learning Research}, 15:\penalty0 1929--1958,
  2014.
\newblock URL \url{http://jmlr.org/papers/v15/srivastava14a.html}.

\bibitem[Szegedy et~al.(2014)Szegedy, Liu, Jia, Sermanet, Reed, Anguelov,
  Erhan, Vanhoucke, and Rabinovich]{Szegedy-et-al-arxiv2014}
Szegedy, Christian, Liu, Wei, Jia, Yangqing, Sermanet, Pierre, Reed, Scott,
  Anguelov, Dragomir, Erhan, Dumitru, Vanhoucke, Vincent, and Rabinovich,
  Andrew.
\newblock Going deeper with convolutions.
\newblock Technical report, arXiv:1409.4842, 2014.

\bibitem[Thrun(1995)]{thrun95a}
Thrun, Sebastian.
\newblock Lifelong learning: {A} case study.
\newblock Technical Report CMU-CS-95-208, School of Computer Science, Carnegie
  Mellon University, Pittsburgh, PA 15213, November 1995.

\bibitem[Tieleman \& Hinton(2012)Tieleman and Hinton]{tieleman2012lecture}
Tieleman, T and Hinton, G.
\newblock Lecture 6.5-rmsprop: Divide the gradient by a running average of its
  recent magnitude.
\newblock \emph{COURSERA: Neural Networks for Machine Learning}, 4, 2012.

\end{thebibliography}
}

\appendix
\section{Choice of Hyperparameters for Fine Tuning {\tt Net2Net} Transformed Models}

{\tt Net2Net} involves training neural networks in a different setting than
usual, so it is natural to wonder whether the hyperparameters must be changed
in some way.
We found that, fortunately, nearly all of the hyperparameters that are
typically used to train a network from scratch may be used to train
a network with {\tt Net2Net}.
The teacher network does not need to have any of its hyperparameters
changed at all.
In our experience, the student network needs only one modification, which
is to use a smaller learning rate than usual.
We find that the initial learning rate for the student network should
be approximately $\frac{1}{10}$ the initial learning rate for the
teacher network.
This makes sense because we can think of the student network as continuing
the training process begun by the teacher network, and typically the
teacher network training process involves shrinking the learning rate
to a smaller value than the initial learning rate.

\section{Related work}

{\em Cascade-correlation}~\citep{Fahlman90} includes a strategy for
learning by enlarging a pre-existing model architecture. This approach
uses a very specific architecture where each new hidden unit receives
input from all previous hidden units. Also, the pre-existing hidden units
do not continue to learn after the latest hidden unit has been added.
A related approach is to add whole new layers
while leaving the lower layers fixed~\citep{gutstein2008knowledge}.
Both these approaches can be considered part of the {\tt Net2Net} family
of operations for rapidly training a new model given a pre-existing one.
However, these variants
require passing through a temporary period of
low performance after a new component has been added to the network
but not yet been adapted. Our function-preserving initializations avoid
that period of low performance.
(In principle we could avoid this period entirely, but in practice we
usually add some noise to the student model in order to break symmetry
more rapidly---this results in a brief period of reduced performance,
but it is a shorter period and the performance is less impaired than
in previous approaches)
Also, these approaches do not allow
pre-existing portions of the model to co-adapt with new portions of the
model to attain greater performance, while our method does.

Some work on knowledge transfer between neural networks is motivated
by a desire to train deeper networks than would otherwise be possible,
by incrementally training deeper networks with knowledge provided
by smaller ones~\citep{Romero-et-al-TR2014,Simonyan2015}.
These works trained convolutional networks with up to 19 weight layers.
Somewhat perplexingly, we were able to train models of considerably
greater depth
without needing to use knowledge transfer, suggesting that deep models
do not pose nearly as much difficulty as previous authors have believed.
This may be due to our use of a very strong baseline:
Inception~\citep{Szegedy-et-al-arxiv2014}
networks with batch normalization~\citep{Ioffe+Szegedy-2015}
trained using RMSProp~\citep{tieleman2012lecture}.
The models we refer to as ``standard size'' have 25 weight layers on
the shortest path from input to output, and 47 weight layers on the
longest path (3 convolution layers + 11 Inception modules, each
Inception module featuring a convolution layer followed by three
forked paths, the shortest of which involves only one more convolution
and the longest of which involves three more).
Rather than using knowledge transfer to make greater depth possible,
our goal is merely to accelerate the training of a new model when
a pre-existing one is available.

Model compression~\citep{bucilua2006model,hinton2015distilling} is a technique
for transferring knowledge from many models to a single model. It serves a
different purpose than our {\tt Net2Net} technique. Model compression aims to
regularize the final model by causing it to learn a function that is similar
to the average function learned by many different models. {\tt Net2Net} aims
to train the final model very rapidly by leveraging the knowledge in a
pre-existing model, but makes no attempt to cause the final model to be more
regularized than if it had been trained from scratch.

Our {\tt Net2DeeperNet} operator involves inserting layers initialized
to represent identity functions into
deep networks. Networks with identity weights at initialization have been
used before by, for example, ~\citet{Socher13parsingwith}, but to our knowledge
that have not previously been used to design function-preserving transformations
of pre-existing neural networks.

The specific operators we propose are both based on the idea of transformations
of a neural network that preserve the function it represents. This is a closely
related concept to the justification for greedy layerwise pretraining of
deep belief networks~\citep{Hinton06}. A deep belief network is a generative
model that draws samples by first drawing a sample from a restricted Boltzmann
machine then using this sample as input to the ancestral sampling process through
a directed sigmoid belief network. This can be transformed into a deep belief
network with one more layer by inserting a new directed layer in between the
RBM and the directed network. If the new directed layer is initialized with
weights from the RBM, then this added layer is equivalent to performing one
extra step of Gibbs sampling in the RBM. It thus does not change the probability
distribution represented by the deep belief network.
Our function-preserving transformations are similar in spirit.
However, our transformations preserve the function represented by a
feed-forward network. Note that DBN layer stacking does not preserve the
function represented by the deterministic upward pass through DBNs that is
commonly used to define a feed-forward neural network.
Our function-preserving transformations are more general in the sense that
they allow adding a layer with any width greater than the layer below it,
while DBN growth is only known to be distribution-preserving for one specific
size of new layer.

\end{document}